\documentclass{article}

\usepackage[final, nonatbib]{neurips_2024}

\usepackage[utf8]{inputenc} 
\usepackage[T1]{fontenc}    
\usepackage{hyperref}       
\usepackage{url}            
\usepackage{booktabs}       
\usepackage{amsfonts}       
\usepackage{nicefrac}       
\usepackage{microtype}      
\usepackage{xcolor}         
\usepackage{amsmath}
\usepackage{amssymb}
\usepackage{amsthm}
\usepackage{mathtools}
\usepackage{tabularx}
\usepackage[ruled,vlined]{algorithm2e}
\usepackage{cleveref}

\theoremstyle{plain}
\newtheorem{theorem}{Theorem}[section]

\theoremstyle{definition}
\newtheorem{definition}[theorem]{Definition}

\theoremstyle{remark}

\newcommand{\eg}{, e.\ g., }
\newcommand{\ie}{, i.\ e., }
\newcommand{\mcB}{\mathcal{B}}
\newcommand{\mcC}{\mathcal{C}}

\newcommand{\mcN}{\mathcal{N}}
\newcommand{\mcP}{\mathcal{P}}
\newcommand{\mcX}{\mathcal{X}}
\creflabelformat{equation}{#2\textup{#1}#3}

\makeatletter
\newcommand{\nosemic}{\renewcommand{\@endalgocfline}{\relax}}
\makeatother

\title{Higher-Order Message Passing for Glycan Representation Learning}

\author{%
  Roman Joeres \\
  Department of Chemistry and Molecular Biology and\\
  Wallenberg Centre for Molecular and Translational Medicine\\
  University of Gothenburg\\
  Gothenburg, Sweden \\
  and\\
  Saarland Informatics Campus\\
  Saarland University\\
  Saarbruecken, Germany\\
  \AND
  Daniel Bojar \\
  Department of Chemistry and Molecular Biology and\\
  Wallenberg Centre for Molecular and Translational Medicine\\
  University of Gothenburg\\
  Gothenburg, Sweden \\
  \texttt{daniel.bojar@gu.se} \\
}

\begin{document}

\maketitle

\begin{abstract}
  Glycans are the most complex biological sequence, with monosaccharides forming extended, non-linear sequences. As post-translational modifications, they modulate protein structure, function, and interactions. Due to their diversity and complexity, predictive models of glycan properties and functions are still insufficient.
  
  Graph Neural Networks (GNNs) are deep learning models designed to process and analyze graph-structured data. These architectures leverage the connectivity and relational information in graphs to learn effective representations of nodes, edges, and entire graphs. Iteratively aggregating information from neighboring nodes, GNNs capture complex patterns within graph data, making them particularly well-suited for tasks such as link prediction or graph classification across domains.
  
  This work presents a new model architecture based on combinatorial complexes and higher-order message passing to extract features from glycan structures into a latent space representation. The architecture is evaluated on an improved GlycanML benchmark suite, establishing a new state-of-the-art performance. We envision that these improvements will spur further advances in computational glycosciences and reveal the roles of glycans in biology.
\end{abstract}

\section{Introduction}

Glycans are complex carbohydrate structures composed of covalently connected monosaccharides forming branching oligosaccharide trees \cite{Varki2016}. Glycans play crucial roles in numerous biological processes, from regulating the immune response to mediating host-pathogen interactions \cite{Pinho2023, Lundstrm2022}. As post-translational modifications, glycans are also key for modulating protein structure, function, and interactions. Especially understanding lectin-glycan interactions has recently shed light onto receptors for SARS-CoV-2 \cite{nguyen2022sialic}, immune checkpoints in cancer \cite{silva2020glycans}, or erythrocyte invasion by malaria-causing parasites \cite{day2024essential}.

Traditional approaches to glycan analysis have relied on hand-crafted features or simple molecular descriptors, which often fail to capture the full structural complexity of these molecules. Recent advances in machine learning, particularly in graph neural networks (GNNs), have shown promise in addressing these limitations \cite{Bojar2022, burkholz2021using}. GNNs can naturally represent the branched, non-linear structure of glycans and learn relevant features directly from unprocessed data.

Despite these advances, existing GNN models for glycan analysis often struggle to simultaneously capture both the atomic-level details and the higher-order structural information of glycans. Models focusing on atomic-level representations \cite{alkuhlani2023gnngly} may miss important topological features, while those operating on a coarser, monosaccharide level \cite{burkholz2021using} may overlook crucial atomic interactions.

To address these challenges, we present GIFFLAR (Glycan Informed Foundational Framework for Learning Abstract Representations), a novel GNN architecture specifically designed for glycan representation learning. GIFFLAR leverages combinatorial complexes \cite{hajij2022topological} to represent glycans at multiple levels of abstraction—atoms, bonds, and monosaccharides—within a single, unified framework. This multi-level representation, combined with higher-order message passing, enables the model to learn rich, hierarchical features that capture both local and global structural information.

We evaluate GIFFLAR on an expanded and curated version of the GlycanML benchmark suite \cite{xu2024glycanml}, encompassing a diverse set of glycan property prediction tasks. Our experiments demonstrate that GIFFLAR consistently outperforms existing methods across all tasks, including traditional machine learning approaches and state-of-the-art GNN models.

The rest of this paper is organized as follows: \Cref{sec:rel_work} discusses related work in glycan analysis and graph neural networks. \Cref{sec:prelim} introduces the theoretical background on combinatorial complexes and higher-order message passing. In \Cref{sec:experiments}, we describe our experimental setup, including data preparation and model architecture. \Cref{sec:eval} presents our evaluation results and ablation studies. Finally, \Cref{sec:end} discusses our findings and outlines limitations and future work directions.

\begin{table}[t]
    \centering
    \begin{tabular}{lrrrr}
        Dataset Name & Task & \# Mono/Glycan & \# Atoms/Glycan & \#train/\#val/\#test \\ \midrule
        Immunogenicity & Binary & 7.47 & 97.54 & 825/230/113\\
        Glycosylation & 3-class & 9.01 & 115.97 & 1,134/317/163 \\
        Tax. Domain & 5-label & 7.13 & 91.43 & 11,378/3,170/1,579 \\
        Tax. Kingdom & 15-label & 7.13 & 91.43 & 11,378/3,170/1,579 \\
        Tax. Phylum & 43-label & 7.13 & 91.43 & 11,378/3,170/1,579 \\
        Tax. Class & 78-label & 7.13 & 91.43 & 11,378/3,170/1,579 \\
        Tax. Order & 170-label & 7.13 & 91.43 & 11,378/3,170/1,579 \\
        Tax. Family & 265-label & 7.13 & 91.43 & 11,378/3,170/1,579 \\
        Tax. Genus & 394-label & 7.13 & 91.43 & 11,378/3,170/1,579 \\
        Tax. Species & 569-label & 7.13 & 91.43 & 11,378/3,170/1,579 \\
        LG Interactions & regression & 6.54 & 83.86 & 388,731/110,885/55,593\\
        \bottomrule
    \end{tabular}
    \caption{Benchmark task description. All values have been extracted from the training partition of the respective data split.}
    \label{tab:dataset_size}
\end{table}

\section{Related Work}\label{sec:rel_work}

\paragraph{GlycanML} In 2024, Xu et al. proposed to use existing benchmark datasets curated by our group to assess the performance of models for glycan property prediction \cite{xu2024glycanml}. This data was mainly taken from the 2021 version of SugarBase \cite{bojar2021deep} and GlyConnect \cite{alocci2018glyconnect} for the glycosylation classification. Also, in 2024, glycowork v1.3 was released \cite{Thoms2021}, and the included data extends the taxonomy datasets beyond the coverage of GlycanML, comprising 40\% new glycans in our extended benchmark datasets. For the presented work, we thus created a dataset comprising eight taxonomy classification tasks from glycowork and two classification tasks for glycosylation and immunogenicity from GlycanML. As our model requires atomic graphs, we filtered the ten datasets for those glycans that could be translated from IUPAC to SMILES using GlyLES \cite{joeres2023glyles}. An overview of some key properties of the filtered datasets is given in \Cref{tab:dataset_size}.

This previous work showed that the Relational Graph Convolutional Network (RGCN) \cite{schlichtkrull2018modeling} was the best on the single-task datasets. Therefore, we implemented this architecture as one of the baselines and the only one using the same heterogeneous graph architecture our new model will use. For comparability, all models in this study were retrained on the same dataset of glycans, using the same data split across all models.

\paragraph{SweetNet} The standard model to predict properties of glycans is SweetNet, which operates on a topological level\ie the tree of monomers \cite{burkholz2021using}. As one of the first glycan-focused geometric deep learning models, it uses a GNN architecture for feature extraction and a simple MLP as the prediction head. Burkholz et al. showed the superiority of their approach over simple baselines. In downstream experiments, they further showed that SweetNet can identify and extract essential features of glycans. In separate work, Lundstr{\o}m et al. used SweetNet as an encoder for glycans in their LectinOracle \cite{lundstrom2022lectinoracle}, a lectin-glycan interaction prediction model, and Kellman et al. used it as an encoder to predict glycosylation potential at a protein glycosite \cite{Kellman2024}.

\paragraph{GNNGLY} After GlyLES was published, Alkuhlani et al. presented GNNGLY, a GCN-based model for glycan property prediction based on atomic structures computed with GlyLES \cite{alkuhlani2023gnngly}. This model was trained on the original taxonomy classification data from SugarBase. They showed improvement over simple baselines and comparable results to SweetNet. However, the study raises several concerns as it is not reproducible in our hands, and SweetNet was not retrained for the initial study. The latter is necessary for comparability, as the initial datasets for training SweetNet and GNNGLY were not the same. Due to structural ambiguities, GlyLES could not convert all IUPAC-condensed strings of this dataset to SMILES, and the usable data for training and evaluating GNNGLY are different from the one used initially for SweetNet. Due to the lack of code and the insufficient architecture description for GNNGLY, we had to implement the model as good as possible.

\paragraph{GLAMOUR} To our knowledge, GLAMOUR is the only published model incorporating atomic and macromolecular (topological) structures of molecules into one model \cite{mohapatra2022chemistry}. Since it applies to any class of multimeric molecules, it has been tested for glycan property prediction. GLAMOUR operates on a topological level like SweetNet, with nodes representing monomeric units and edges representing their covalent connections. The atomic structures of the monomers and their connections are used to compute fingerprints as the initial features of the nodes and edges. GLAMOUR offers five GNN architectures as the back-end of the model. However, we could not apply GAT and GCN to monosaccharides. Among the other three, namely MPNN, Weave, and AttentiveFP, the MPNN performed best (see \Cref{app:ablation}). Therefore, we used the MPNN backend in the GLAMOUR baseline.

\section{Theory}\label{sec:prelim}

This section will explain how we interpret glycans as graphs and use combinatorial complexes and higher-order message passing to compute their topology-aware embeddings.

\paragraph{Combinatorial Complexes} Combinatorial complexes (CCs) are the abstract superclass of simplicial complexes, cellular complexes, and more. As they are central to our work, we will precisely define CCs following the definition of Eitan et al. \cite{eitan2024topological}. Definitions of simplicial and cellular complexes are given in \cite{bodnar2021weisfeilerC, bodnar2021weisfeilerT}.

\begin{definition}
    A combinatorial complex (CC) is a tuple $C := (S, \mathcal{X}, f)$ where $S$ is a set, $\mcX\subseteq\mcP(S)\setminus\emptyset$ is a subset of its powerset $\mcP(S)$, and $f\coloneqq \mathcal{X}\mapsto\mathbb{Z}_0^+$ is a function with the following properties:
    \begin{align}
        &\forall s\in S.\ \{s\}\in \mcX\\
        &\forall x, y\in \mcX.\ x\subset y\implies f(x)<f(y)\text{\ie $f$ is order-preserving.}
    \end{align}
\end{definition}

The elements of $S$ correspond to a graph's nodes (or vertices); elements of $\mcX$ are called cells. Different cell ranks are based on the organization level within the combinatorial complex. In a general CC, all cells of rank $i$ form the $i$-skeleton, denoted as $\mcX_i$. For our work, nodes are 0-cells (representing atoms), edges between nodes are 1-cells (representing bonds), and 2-cells are collections of edges (representing monomers).

\begin{figure}[t]
    \centering
    \includegraphics[width=\linewidth]{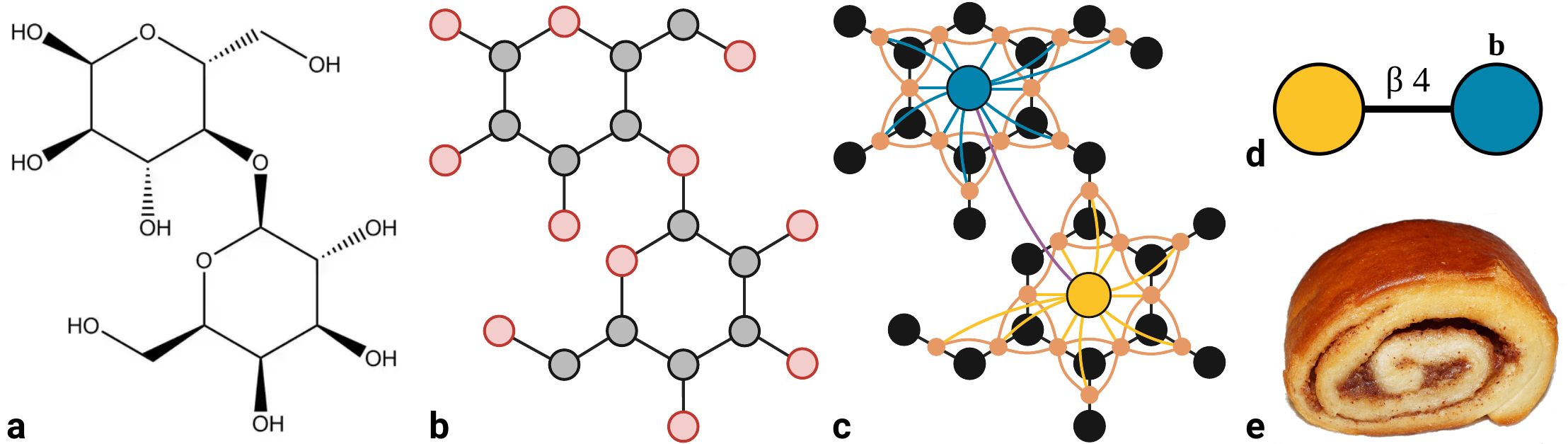}
    \caption{Lactose in four abstractions. Panel a shows the chemical formula, and Panel b shows the graph-of-atoms (GOA). Panel c shows the combinatorial complex and cells of all ranks. 0-cells in black represent the atoms; 1-cells in orange are bonds between atoms, and 2-cells in yellow and blue denote whole monosaccharides as defined by the GlycoDraw depiction in panel d \cite{lundstrom2023glycodraw}. Panel e shows an inspirational and motivational Gifflar, a Swedish cinnamon roll.}
    \label{fig:cc}
\end{figure}

\paragraph{Neighborhood functions} Within and between ranks $C$, one can define a neighborhood function $\mcN\coloneqq\mcX\mapsto\mcP(\mcX)$. For $x\in\mcX_i$ and $j>i$, the following neighborhood functions are relevant to our work:
\begin{align}
    \mcB_{i,j}(x) &= \{y\in\mcX_i\mid\exists z\in\mcX_j\text{ s.t. }x,y\subseteq z\} \label{eq:nB}\\
    \mcN^\downarrow_{i,j}(x) &= \{y\in\mcX_j\mid x\subseteq y\} \label{eq:nD}
\end{align}
where $i$ and $j$ are different ranks of a CC. \Cref{eq:nB} defines neighborhoods within rank $i$ over shared cells with higher ranks. Conversely, \Cref{eq:nD} defines neighborhoods between ranks, where \Cref{eq:nD} is the set of higher-rank neighbors. A more general definition is given in \cite{hajij2022topological}. \Cref{fig:cc} visualizes all definitions and relations on the example of lactose.

\paragraph{Message passing neural networks} To compute embeddings of graph-structured data, such as molecules or networks, graph neural networks (GNNs) are the most powerful tool. While GNNs comprise many architectures, such as graph convolution and graph attention, we focus on message-passing neural networks (MPNNs). The first step is to compute embeddings $\mathbf{h}_x^l$ for nodes $x$ in layers $l\in[L]$ based on their neighbors $y\in\mcN(x)$ and the edge data $\mathbf{e}_{x,y}$ as defined in 

\begin{equation}
    \mathbf{h}_{x}^{l+1} = \sigma \left( \mathbf{h}_x^l, \bigoplus \limits_{y\in\mcN(x)} \theta^l \left( \mathbf{h}_x^l, \mathbf{h}_y^l, \mathbf{e}_{x, y} \right) \right) \label{eq:nagg}
\end{equation}

with $\sigma$ being a non-linear activation function, $\theta^l$ as layer-specific, differentiable functions, and $\bigoplus$ is a permutation-invariant aggregation function. Then, the node embeddings $\mathbf{h}_x^L$ of the final layer $L$ are aggregated into a graph embedding $\mathbf{h}^L$, following

\begin{equation}
    \mathbf{h}^L = \sigma\left(\bigoplus \limits_{x\in\mcC} \theta \left( \mathbf{h}_x^L \right) \right) \label{eq:nro}
\end{equation}
which represents the whole graph structure \cite{gilmer2017neural}.

\paragraph{Higher-order message passing} Using the neighborhoods on CCs defined above, we can define a message-passing scheme on CCs following Hajij et al. \cite{hajij2022topological}.

\begin{definition}
    Let $\mcN=\{\mcN_1,\dots,\mcN_n\}$ be a set of neighborhood functions defined on $\mcC$\eg $\mcN=\{\mcB_{0,1}, \mcN^\downarrow_{0,1}\}$. Furthermore, $\mathbf{h}_x^l$ is the embedding of cell $x\in\mcX$ in layer $l\in\{0,\dots, L\}$. Higher-order message passing on $C$ is defined following this update rule:
    \begin{align}
        \mathbf{h}_{x}^{l+1} = \sigma \left( \mathbf{h}_x^l, \bigotimes\limits_{\mcN_k\in\mcN} \bigoplus\limits_{y\in\mcN_k} \theta_{\mcN_k}^l \left( \mathbf{h}_x^l, \mathbf{h}_y^l \right) \right) \label{eq:homp}
    \end{align}
\end{definition}
Here, $\theta^l_{\mcN_k}$ is a layer-specific and neighborhood-specific differentiable function\eg neural networks, and $\bigotimes$ and $\bigoplus$ are permutation-invariant aggregation functions gathering inter-neighborhoods and intra-neighborhoods, respectively.

To represent $C$ in a single embedding, the readout function is defined as follows:
\begin{align}
    \mathbf{h}^L = \sigma\left(\bigotimes\limits_{r=0}^R\bigoplus\limits_{x\in\mcC}\theta_r\left(\mathbf{h}_x^L\right)\right)
\end{align}
with $R$ being the number of ranks in $\mcX$ and $L$ the number of layers to compute. Everything else is as defined for \Cref{eq:homp}.

\section{Experiment Setup}\label{sec:experiments}

In total, we trained nine property-prediction models on four representations of glycans. Simple baselines, namely Random Forests \cite{breiman2001random}, Support Vector Machines \cite{boser1992training}, Gradient Boosting \cite{friedman2001greedy}, and Multilayer Perceptrons \cite{werbos1974beyond}, on atom-level Morgan Fingerprints\cite{rogers2010extended}, GNNGLY on atom-level homogeneous graphs, SweetNet and GLAMOUR on topology-level homogeneous graphs, and RGCN and GIFFLAR on heterogeneous graphs. These have three node types: atoms, bonds, and monosaccharides, corresponding to 0-cells, 1-cells, and 2-cells, respectively, as detailed in \Cref{sec:prelim}. They are connected based on neighborhoods as defined in \Cref{eq:nB,eq:nD}, for intra-rank and inter-rank neighborhoods, respectively. Because all models except SweetNet require atomic graphs as input, we translated all IUPAC-condensed notations to SMILES using GlyLES \cite{joeres2023glyles} and filtered for those glycans for which GlyLES could compute SMILES strings. We then computed atomic graphs from these SMILES strings using RDKit. 

The data was split randomly, and all models were trained and evaluated on the same partitions (see \Cref{app:train_det} for more details). All performance reported in this work was computed on the validation set. These performances were used to compare models and justify architectural decisions. Every model is biased toward the data that influenced its development \cite{kaufman2012leakage}. Therefore, we used a separate holdout dataset to report the state-of-the-art performance on unseen data for the best model chosen on the validation set.

Similar to GlycanML, we trained on the Immunology and Glycosylation datasets as binary and three-class, single-label classification tasks, respectively. In Alkuhlani et al.\cite{alkuhlani2023gnngly} and Xu et al. \cite{xu2024glycanml}, training taxonomy classification as a multi-class, single-label problem led to accuracy measurements that are hard to interpret because a perfect model would not achieve an accuracy of 1, since the same glycan may be conserved across multiple taxonomic groups. Therefore, we changed the task for taxonomy predictions to a multi-label classification. This also reduced the dataset sizes, as each glycan was now only present once with all its labels instead of once per label. Lastly, we trained different encoder combinations on the lectin-glycan interaction dataset from glycowork. The task was to predict the z-score normalized relative fluorescence values extracted from glycan array experiments. Again, we removed those glycans GlyLES could not translate to SMILES strings.

\subsection{Model Architecture}

For GIFFLAR, we used an architecture similar to Graph Isomorphism Networks \cite{xu2018powerful} and instantiated \Cref{eq:homp} as 
\begin{align}
    \mathbf{h}_x^{l+1} &= \sum_{\mcN_k\in\mcN} \theta_{\mcN_k}^l \left( (1 + \epsilon) \cdot \mathbf{h}_x^l + \sum\limits_{y\in\mcN_k} \mathbf{h}_y^l\right),\text{ with}\\
    \mcN &= \{\mcB_{0,0},\ \mcB_{1,1},\ \mcB_{2,2},\ \mcN^\downarrow_{0,1},\ \mcN^\downarrow_{1,2}\}
\end{align}
and $\theta_k^l$ consisting of a fully connected layer, a parameterized ReLU activation function, dropout with $p=0.2$, and batch normalization. This architecture is shared across layers $l$ and neighborhoods $\mcN_k$, but not the weights. The readout for the final graph embedding is a simple mean over all nodes. We explored other approaches, which did not result in higher performance (see \Cref{app:model_comp}). The classification head is a simple, two-layered MLP with a parameterized ReLU activation function and a dropout layer ($p=0.2$). The final model takes 128-dimensional node features as input. We use fixed, random embeddings per node class, e.g., C or O for atom nodes. The feature vectors are scaled up to 1024-dimensional embeddings in the hidden layers. As a result, the model has 35.1M trainable parameters.

For the prediction of the lectin-glycan interactions, we pair the glycan encoder with a protein encoder to embed the lectins. We tested four protein language models (PLMs), namely ESM-2 t33\cite{lin2022language}, Ankh-base\cite{elnaggar2023ankh}, ProtBERT\cite{elnaggar2021prottrans}, and ProstT5\cite{heinzinger2024sty}. For the prediction model, we used the protein encoders to compute fixed protein embeddings and trained GIFFLAR or SweetNet from scratch.

\subsection{Positional Encodings}
In graphs, node features can be computed based on their connectivity and neighborhoods within the graph. Therefore, one can compute so-called \textit{Positional Encodings}. In this work, we experimented with two types of PEs: RandomWalk PEs and Laplacian PEs. $k$-dimensional RandomWalk PEs are computed from a $k$-step random walk through the graph, starting from each node. After each step $i$, the number of walkers in a node is extracted as the $i$-th feature for the positional encoding \cite{dwivedi2021graph}. $k$-dimensional Laplacian PEs are defined as the first $k$ dimensions of the eigenvectors of the graph's Laplacian with randomly assigned signs \cite{dwivedi2023benchmarking}.

\section{Evaluation}\label{sec:eval}

Combinatorial complexes present a natural choice to represent glycans as graph objects, combining their atomic representation and topological structure. 
We evaluated the performance of the models using accuracy, Area Under the Receiver Operating Characteristics curve (AUROC), and Matthews Correlation Coefficient (MCC) on all ten datasets. The three metrics cannot be immediately aggregated and compared across all datasets, as a gain of .1 in accuracy differs from gaining .1 in AUROC, and the difficulty of such gains varies between datasets. Therefore, we applied min-max normalization to all metrics to bring them to a comparable scale. This resulted in 30 performance metrics per model ($10$ datasets $\times\ 3$ metrics). We then combined the scores (see \Cref{algo:anp}), as no model was strictly best in all 30 metrics. This allowed us to assign a scalar performance metric to each model as the sum of its 30 normalized performances to identify the best model on average.

\begin{algorithm}[ht]
	\SetAlgoLined\nosemic
	\KwIn{Performance Tensor $P\in\mathbb{R}^{\mathcal{M}\times D\times M}$}
        $\mathcal{M}$ is the number of metrics, $D$ is the number of datasets, and $M$ is the number of models.\;
	\For{$m$ \textbf{in} $\mathcal{M}$}{
		\For{$d$ \textbf{in} $D$}{
            $d_{min}\leftarrow \min P[m, d, :]$\;
            $d_{max}\leftarrow \max P[m, d, :]$\;
            $P[m, d, :]\leftarrow \frac{P[m, d, :] - d_{min}}{d_{max} - d_{min}}$\;
		}
	}
    $output\leftarrow\left\lbrack\sum P[:, :, m] \textbf{ for }m\textbf{ in }M\right\rbrack$\;
    \Return{output}
	\caption{Computation of Accumulated, Normalized Performance}
    \label{algo:anp}
\end{algorithm}

\begin{figure}[t]
    \centering
    \includegraphics[width=\linewidth]{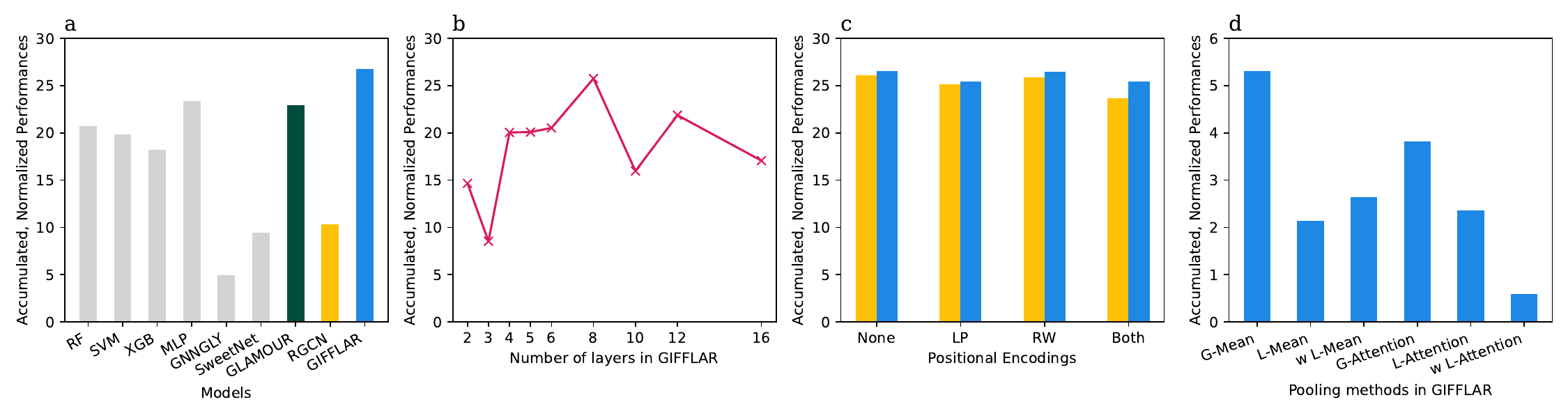}
    \caption{\textbf{a} Averaged normalized performances (ANPs) comparing GIFFLAR (blue) to the eight baselines. \textbf{b} Performance on different depths of GIFFLAR. \textbf{c} Comparison of different combinations of positional encodings and GIFFLAR with 128 feature dimensions (blue) to 1024 (red). \textbf{d} Comparison of different pooling mechanisms on GIFFLAR.}
    \label{fig:ablation}
\end{figure}

We compared the final model to retrained baselines based on Morgan 1024-bit Fingerprints (Random Forests, Support Vector Machines, GradientBoosting, and an MLP), homogeneous graphs (GNNGLY on an atomic level, and SweetNet and GLAMOUR on a topological level), and RGCN on heterogeneous graphs. Here, again, our new GIFFLAR architecture showed superior performance, being the best-in-class model despite a lower number of trainable parameters (see \Cref{fig:ablation}a and \Cref{tab:model_prop,tab:comp_mcc}). The fingerprint-based baselines appear very strong, which is a common phenomenon observed in the field \cite{jiang2021could}.

\begin{table}[t]
    \centering
    \begin{tabular}{ccccccccccc}\toprule
        Frac. & Immun. & Glycos. & D & K & P & C & O & F & G & S\\
        \midrule
        full & .8145 & .9655 & .9531 & .9214 & .8481 & .7877 & .6464 & .6236 & .5381 & .4943 \\
        OOD & 1 & .6902 & .9247 & .9198 & .7952 & .7234 & .5507 & .5192 & .4381 & .4043 \\
        \bottomrule
    \end{tabular}
    \caption{Matthews Correlation Coefficient of the final GIFFLAR architecture on the test set, establishing a new SOTA. Performances on the full test set and on only out-of-distribution (OOD) glycans are shown. 
    }
    \label{tab:test_mcc}
\end{table}

\Cref{tab:test_mcc} shows the performance of GIFFLAR on the held-out test set\ie data that did not influence the model development. The performance on out-of-distribution (OOD) data may vary from this, as shown in \cite{joeres2023datasail}. Therefore, we report the performance of the OOD fraction of the test set separately. 
For this work, we defined OOD glycans as glycans with less then 0.75 Tanimoto similarity between their monosaccharide-based fingerprints.
Notably, the variance of the OOD-MCC is relatively high for the Immunogenicity and Glycosylation datasets, as the test set only contained 31 and 12 OOD samples, respectively. The test sets of the taxonomy datasets contained over 400 OOD samples and are, therefore, more stable.

\begin{figure}[t]
    \centering
    \includegraphics[width=\linewidth]{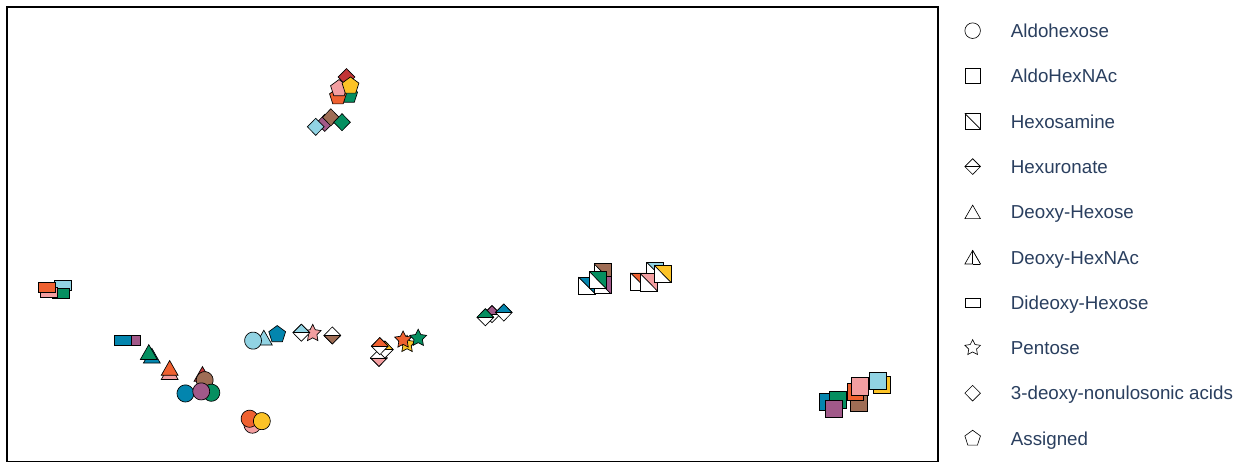}
    \caption{t-SNE plot of monosaccharide embeddings of GIFFLAR.}
    \label{fig:monosacc}
\end{figure}

\Cref{fig:monosacc} visualizes GIFFLAR embeddings of different monosaccharides. This provides insights into how GIFFLAR extracts features from structures and puts them in relation to each other. The Symbol Nomenclature For Glycans (SNFG) organizes monosaccharides via colors and shapes that hint at the chemical composition of the monosaccharide (e.g., circles being hexoses, or blue monosaccharides sharing the same underlying stereochemistry), which explains some of the resulting clustering. Yet it is interesting to note that monosaccharides do not solely cluster according to this chemical similarity, indicating that task-specific properties (predicting the taxonomic Domain in which a glycan is expressed) have been learned here, such as the classic animal uronic acids (GlcA, IdoA (blue and brown divided diamonds, respectively)) clustering distinctly from uronic acids that may only appear in bacteria (AltA (purple divided diamond) etc.).

\begin{figure}[t]
    \centering
    \includegraphics[width=\linewidth]{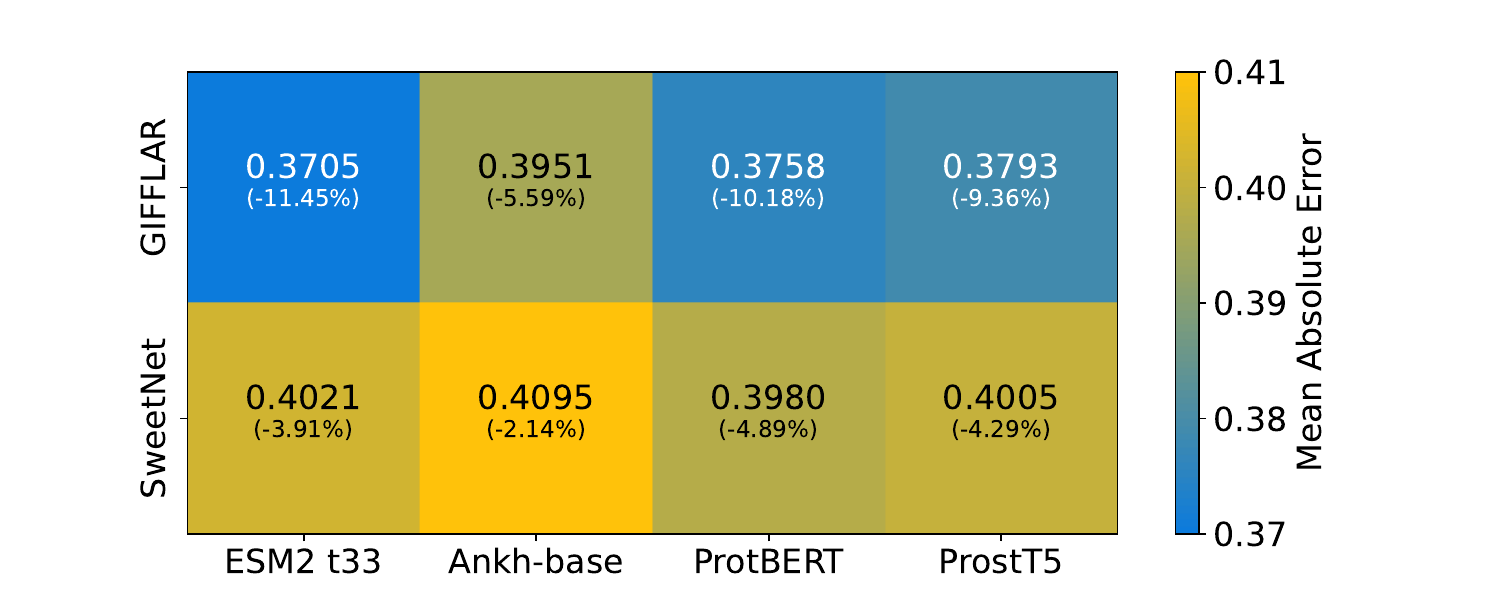}
    \caption{Mean Absolute Error of different glycan and protein encoders. In brackets, we denote the relative improvement compared to the LectinOracle baseline ($\text{MAE} = 0.4285$).}
    \label{fig:lgi}
\end{figure}

In \Cref{fig:lgi}, we compare the performance of GIFFLAR to SweetNet as the glycan encoder for lectin-glycan interaction prediction. The baseline model is the LectinOracle combining SweetNet and ESM-1b \cite{lundstrom2022lectinoracle}. Here, we report the improvement of MAE over this baseline for 8 combinations of GIFFLAR and SweetNet with one of the protein language models ESM-2, Ankh, ProtBERT, and ProstT5. We can see that GIFFLAR clearly outperforms SweetNet as a glycan encoder and improves the prediction quality by more than 10\% when paired with ESM2-t33.

\subsection{Ablation Studies}\label{app:ablation}

In early experiments, we investigated the influence of model depth on performance. We found that an 8-layered GIFFLAR model performed best, on average, on property prediction tasks (see \Cref{fig:ablation}b). We also investigated how node feature dimensions and positional encodings (PEs) impacted model performance. We tested GIFFLAR with 128-dimensional and 1024-dimensional random features combined with RandomWalk PEs and Laplacian PEs. Neither individual nor combined PEs consistently improved the performance of the GIFFLAR model (see \Cref{fig:ablation}c). Further, we observed almost equal performance of 128- and 1024-dimensional feature vectors. Following the principle of Occam's Razor, we thus favored 128-dimensional features, which were then scaled up to 1024 dimensions in the first GNN layer.

\Cref{fig:ablation}d compares different pooling mechanisms for the GIFFLAR model with 128-dimensional feature vectors. Here, we only compared the models on the Glycosylation, Immunogenicity, and Domain datasets to reduce runtime. The six modes we compared were:
\begin{itemize}
    \item \textbf{Global Mean}: Computing the mean over all embeddings in layer $L$ regardless of their cell rank.
    \item \textbf{Local Mean}: First, compute the mean over all embeddings in Layer $L$ within each cell rank and then take the mean over the three rank means.
    \item \textbf{weighted Local Mean}: Same as Local Mean, but multiply each cell rank with a learnable weight before summation.
    \item \textbf{Global Attention}: Computing the Soft Attention over all embeddings in layer $L$ regardless of their cell rank.
    \item \textbf{Local Attention}: First, compute the Soft Attention over all embeddings in Layer $L$ within each cell rank and then take the mean over the three rank means.
    \item \textbf{weighted Local Attention}: Same as Local Attention, but multiply each cell rank with a learnable weight before summation.
\end{itemize}

Surprisingly, attention-free pooling functions generally performed better than attention-based ones in this analysis. Further, global pooling was better than local or weighted local methods. We conclude that global mean pooling seems to constitute the most performant pooling operation for GIFFLAR.

We also conducted ablation studies to determine the backbone structure of GLAMOUR and the influence of PEs on the performance of RGCN. The results are described in \Cref{app:model_comp}.

\section{Discussion, Conclusion, Limitations, and Future Work}\label{sec:end}

Here, we presented GIFFLAR, a novel graph neural network architecture designed specifically for glycan representation learning. By leveraging combinatorial complexes and higher-order message passing, GIFFLAR achieves state-of-the-art performance across a wide range of glycan property prediction tasks.
GIFFLAR consistently outperforms existing methods, including traditional machine learning approaches using Morgan fingerprints, homogeneous graph neural networks such as GNNGLY, SweetNet, and GLAMOUR, and heterogeneous graph models such as RGCN. The superior performance of GIFFLAR can likely be attributed to its ability to capture both atomic-level structure and topological information of glycans simultaneously. Due to its superior performance, we expect that GIFFLAR will become the new foundation in computational glycobiology and yield improvements in relevant downstream tasks. Another advantage of GIFFLAR over the baselines is that the feature extractor can be trained end-to-end specifically for a task\eg in lectin-glycan interaction prediction.

Using combinatorial complexes allows GIFFLAR to represent glycans at multiple levels of abstraction—atoms, bonds, and monosaccharides—in a single, unified framework. This multi-level representation, combined with higher-order message passing, enables the model to learn rich, hierarchical features crucial for accurate glycan property prediction. While we focused on glycans in this work, the principles behind GIFFLAR could be extended and adapted to other complex biomolecules, such as metabolites or lipids. Exploring these applications could further broaden the impact of our approach.

GIFFLAR achieves high predictive performance, yet interpreting its learned representations remains challenging. Developing techniques to visualize and explain model predictions would be valuable for glycobiologists and could lead to new insights into glycan structure-function relationships. We further anticipate that future work could explore the value of pre-training such models on larger sets of unlabeled data, similar to established procedures in protein representation learning.

In conclusion, GIFFLAR presents a significant advance in glycan representation learning, demonstrating the power of combining combinatorial complexes with higher-order message passing in graph neural networks. As glycomics continues to grow in importance within the life sciences, we envision that GIFFLAR and its future extensions will play a crucial role in unlocking the full potential of glycan-related research and applications. A prime example application of GIFFLAR as a trainable feature extractor for glycans would be for instance the field of lectin-glycan interaction prediction.

\begin{ack}
R.J. was supported by the Knut and Alice Wallenberg Foundation and the University of Gothenburg. This work was funded by a Branco Weiss Fellowship - Society in Science awarded to D.B., the Knut and Alice Wallenberg Foundation, and the University of Gothenburg, Sweden.

The authors thank the Drug Bioinformatics group at the Helmholtz Institute for Pharmaceutical Research Saarland, Saarbrücken, Germany, for providing the computing resources to conduct the experiments.
\end{ack}

\section*{Code Availability}
The implementation of GIFFLAR is publicly available at \url{github.com/BojarLab/GIFFLAR}.

\section*{Data Availability}
The data (including the data splitting) used in this study is publicly available in the \verb|datasets| folder of \url{github.com/BojarLab/GIFFLAR}.

\section*{Author Contributions}
R.J. implemented the code and conducted the experiments. R.J. and D.B. wrote the manuscript.

\section*{Competing Interests}
R.J. has no competing interests. D.B. consults for SweetSense Analytics AB.

\bibliographystyle{unsrt}
\bibliography{references}

\newpage
\appendix

\setcounter{figure}{0}
\setcounter{table}{0}
\renewcommand{\thefigure}{A\arabic{figure}}
\renewcommand{\thetable}{A\arabic{table}}

\section{Training Details}\label{app:train_det}

\Cref{tab:model_prop} lists core measurements of the models we compare. We used the Domain prediction from the taxonomy collection as an example dataset. The table's average number of nodes per graph is computed on the training set. The first three models (RF, SVM, and XGB) were trained using scikit-learn v1.5.1 \cite{scikitlearn} on a regular CPU. The deep learning models were trained using PyTorch v2.3.1 \cite{paszke2019pytorch}, PyTorch Geometric v2.5.3 \cite{fey2019fast}, and PyTorch Lightning v2.3.2 \cite{Falcon_PyTorch_Lightning_2019} on an NVIDIA GeForce RTX 3090 with 24GB GPU RAM. All metrics for all models were computed using TorchMetrics v1.4.0 \cite{detlefsen2022torchmetrics}.
 
\begin{table}[h]
    \centering
    \begin{tabularx}{\linewidth}{l>{\raggedleft\arraybackslash}X>{\raggedleft\arraybackslash}X>{\raggedleft\arraybackslash}X}
        \toprule
        Model & Training time & $\varnothing$ \# nodes per graph & \# param. \\
        \midrule
        \multicolumn{4}{c}{Morgan Fingerprint (1024-bit)}\\
        RF & 0.1 min & -- & -- \\
        SVM & 2.2 min & -- & -- \\
        XGB & 1.7 min & -- & -- \\
        MLP & 11.7 min & -- & 527K \\
        \toprule
        \multicolumn{4}{c}{Homogeneous Graphs}\\
        GNNGLY & 20.5 min & 91 & 8M \\
        SweetNet & 22.7 min & 13 & 37.2M \\
        GLAMOUR & 9.5 min & 7 & 18.2M \\
        \toprule
        \multicolumn{4}{c}{Heterogeneous Graphs}\\
        $\text{RGCN}_\text{RW}$ & 55.3 min & 195 & 41.9M \\
        GIFFLAR & 45.0 min & 195 & 35.1M \\
        \bottomrule
    \end{tabularx}
    \caption{Measurements of models on Domain Dataset (as an exemplary dataset).}
    \label{tab:model_prop}
\end{table}

\section{Further Model Analysis}\label{app:model_comp}

\begin{figure}
    \centering
    \includegraphics[width=\linewidth]{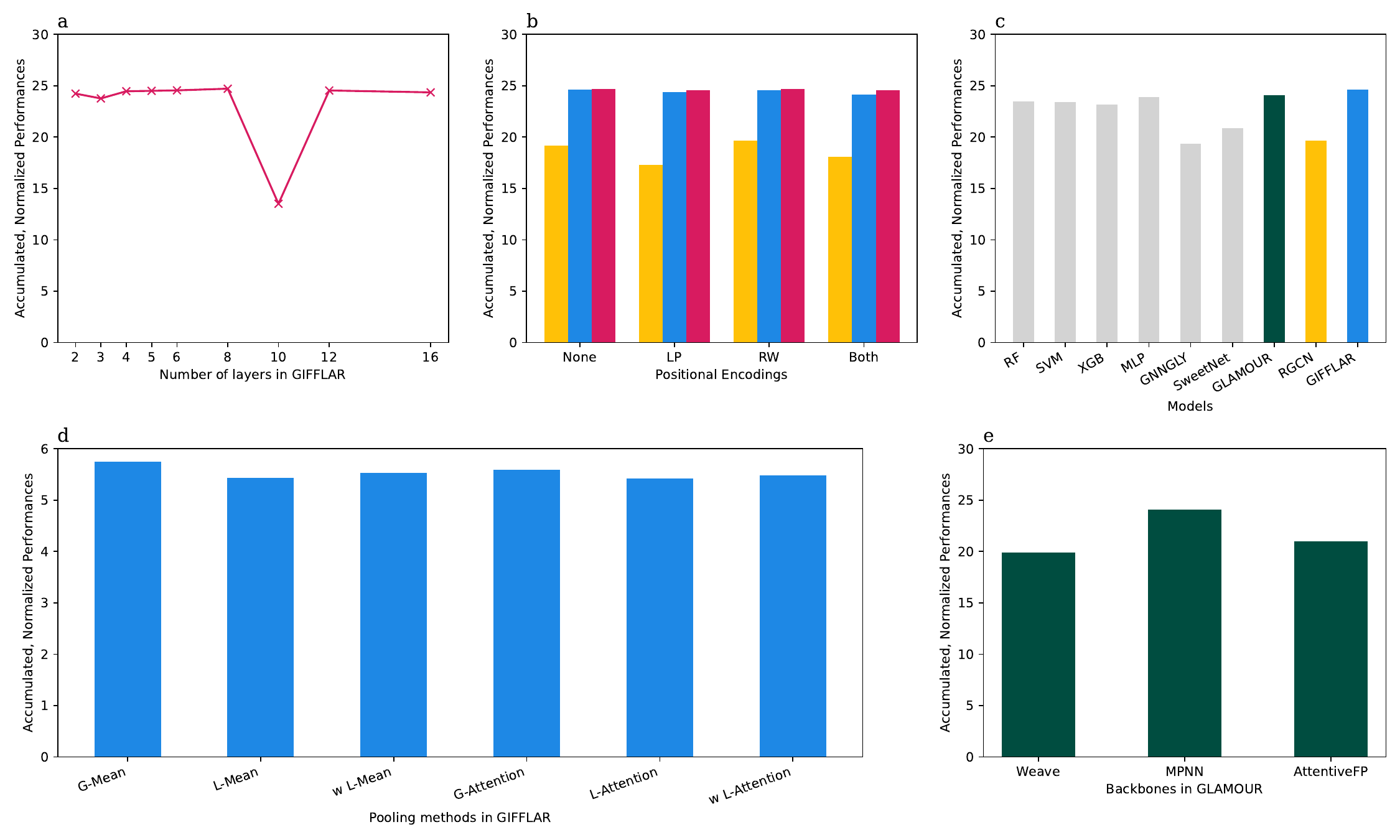}
    \caption{\textbf{a-d} Reevaluation of \Cref{fig:ablation} without the normalization in \Cref{algo:anp}. \textbf{b} contains additional results for testing PEs for the RGCN model. \textbf{e} Comparison of three backbones for the GLAMOUR model.}
    \label{fig:anp_raw}
\end{figure}

To investigate the impact of Algorithm 1 on the performance comparison in \Cref{fig:ablation}, we report the same plot but unnormalized in \Cref{fig:anp_raw} (except for MCC, which was transformed by $x\mapsto\frac{x+1}{2}$ to be in the interval \verb|[0,1]|). Compared to \Cref{fig:ablation}, the ordering of the models by performance does not change significantly, but the differences between the models shrink.

We report Matthews Correlation Coefficients (MCC) for the final models and baselines to give more perspective on the model performances (see \Cref{tab:comp_mcc}). For the multi-label datasets, the MCC was computed per label and then averaged over the labels unweighted.

As Mohapatra et al. do not name one backbone architecture as the best among the five provided \cite{mohapatra2022chemistry}, we compared the usable ones. The provided implementation does not allow for applying GCN— and GAT-based models to graphs representing monomeric molecules. In \Cref{fig:anp_raw}a, we compared the Weave, MPNN, and AttentiveFP models, among which the MPNN performed best and was selected as the GLAMOUR backbone for the baseline model in this work.

As the RGCN model is the only baseline without specified featurization of the nodes, we tested PEs here as well and found that adding positional encodings from random walks improved the performance as shown in \Cref{fig:anp_raw}b.

\begin{table}[t]
    \centering
    \begin{tabularx}{\linewidth}{lcccccccccc}\toprule
        Model & Immun. & Glycos. & D & K & P & C & O & F & G & S\\
        \midrule
        \multicolumn{11}{c}{Morgan Fingerprint (1024-bit)}\\
        RF & .8223 & .9648 & .9129 & .8749 & .8010 & .7094 & .5546 & .4944 & .4613 & .4439 \\
        SVM & .8034 & .9648 & .8793 & .8403 & .7466 & .6398 & .4588 & .4369 & .4137 & .3880 \\
        XGB & .8302 & .9824 & .8718 & .8348 & .7393 & .6282 & .4705 & .4420 & .3870 & .3467 \\
        MLP & .8481 & .8704 & .9106 & .8763 & .8033 & .7206 & .5413 & .5097 & .4708 & .4282 \\
        \toprule
        \multicolumn{11}{c}{Homogeneous Graphs}\\
        GNNGLY & .5328 & .7611 & .7717 & .7747 & .6609 & .4685 & .0156 & .0150 & .0151 & .0154 \\
        SweetNet & .7590 & .8784 & .8841 & .7704 & .6232 & .5288 & .0156 & .1872 & .0151 & .1175 \\
        GLAMOUR & \textbf{.9212} & .9767 & .9111 & .8704 & .7864 & .6857 & .4998 & .4785 & .4320 & .4407 \\
        \toprule
        \multicolumn{11}{c}{Heterogeneous Graphs}\\
        RGCN & .6954 & .0000 & .8810 & .8409 & .7211 & .4039 & .2288 & .0314 & .2530 & .0194 \\
        GIFFLAR & .8930 & \textbf{.9883} & \textbf{.9298} & \textbf{.9011} & \textbf{.8278} & \textbf{.7714} & \textbf{.6118} & \textbf{.5795} & \textbf{.5391} & \textbf{.4898} \\
        \bottomrule
    \end{tabularx}
    \caption{Comparison of Matthews Correlation Coefficient of GIFFLAR to the seven baselines. Bold values mark the best performances on a certain dataset. The single-letter column names refer to the taxonomic levels of D--domain, K--kingdom, P--phylum, C--class, O--order, F--family, G--genus, and S--species. We will also use these in all other tables.}
    \label{tab:comp_mcc}
\end{table}

\end{document}